%% file: main.tex
\def\year{2022}\relax
\documentclass[letterpaper]{article} 
\usepackage{aaai22}  
\usepackage{times}  
\usepackage{helvet}  
\usepackage{courier}  
\usepackage[hyphens]{url}  
\usepackage{graphicx} 
\def\UrlFont{\rm}  
\usepackage{natbib}  
\usepackage{caption} 
\usepackage{booktabs}       
\usepackage{amsfonts}       

\DeclareCaptionStyle{ruled}{labelfont=normalfont,labelsep=colon,strut=off} 
\usepackage{algorithm}
\usepackage{algorithmic}
\newtheorem{theorem}{Theorem}

\newtheorem{prop}[theorem]{Proposition}

\usepackage{url}            
\usepackage{nicefrac}       
\usepackage{microtype}      
\usepackage{newfloat}
\usepackage{listings}
\usepackage{wrapfig}

\floatstyle{ruled}
\newfloat{listing}{tb}{lst}{}
\floatname{listing}{Listing}
\title{SPGNN:  Recognizing Salient Subgraph Patterns \\ via Enhanced Graph Convolution and Pooling
}

\author {
    Zehao Dong,\textsuperscript{\rm 1}
    Muhan Zhang \textsuperscript{\rm 2}
    Yixin Chen\textsuperscript{\rm 1} \\
}
\affiliations {
    \textsuperscript{\rm 1} Department of Computer Science and Engineering, Washington University in St. Louis \\
    \textsuperscript{\rm 2}  Institute for Artificial Intelligence, Peking University \\
}

\begin{document}
\maketitle

\begin{abstract}
    Graph neural networks (GNNs) have revolutionized the field of machine learning on non-Euclidean data such as graphs and networks. GNNs effectively implement node representation learning through neighborhood aggregation and achieve impressive results in many graph-related tasks. However, most neighborhood aggregation approaches are summation-based, which can be problematic as they may not be sufficiently expressive to encode informative graph structures. Furthermore, though the graph pooling module is also of vital importance for graph learning, especially for the task of graph classification, research on graph down-sampling mechanisms is rather limited. 
   
   To address the above challenges, we propose a concatenation-based graph convolution mechanism that injectively updates node representations to maximize the discriminative power in distinguishing non-isomorphic subgraphs. In addition, we design a novel graph pooling module, called WL-SortPool, to learn important subgraph patterns in a deep-learning manner. WL-SortPool layer-wise sorts node representations (i.e. continuous WL colors) to separately learn the relative importance of subtrees with different depths for the purpose of classification, thus better characterizing the complex graph topology and rich information encoded in the graph. We propose a novel Subgraph Pattern GNN (SPGNN) architecture that incorporates these enhancements. We test the proposed SPGNN architecture on many graph classification benchmarks. Experimental results show that our method can achieve highly competitive results with state-of-the-art graph kernels and other GNN approaches.
\end{abstract}

\section{Introduction}
\label{sec: intro} 
\input{1_intro}

\section{Related Works}
\label{sec: relate} 
\input{2_related}

\section{Subgraph Pattern Graph Neural Network (SPGNN)}
\label{sec: method} 
\input{3_settings}

\section{Experimental Results}
\label{sec: exp result} 
\input{4_evaluation}

\section{Conclusions}
\label{sec: iconc} 
\input{5_conclusion}

\bibliography{reference}

\onecolumn
\appendix
\input{6_supplementary}

\end{document}

%% file: 1_intro.tex
In recent years, there has been a growing interest to develop graph neural networks (GNNs) for learning from graph-structured data, and have revolutionalized machine learning methodology in various field including social network analysis\cite{monti2017geometric,ying2018graph} \cite{monti2017geometric,ying2018graph}, bioinformatics \cite{fout2017protein,zitnik2018modeling, dong2023highly, dong2024large}, analog circuit design \cite{cao2022domain,dong2022pace} etc. 
To learn from these graphs, various differentiable graph neural network (GNN) frameworks have been proposed to capture graph structures, and graph convolution networks inspired by the successful application of convolutional neural networks (CNNs) in computer vision play a fundamental role in these graph deep learning architectures. 

As some of the earliest works, spectral-based GNNs 
\cite{bruna2013spectral, defferrard2016convolutional} define graph convolution operations from the perspective of graph signaling processing. However, spectral GNNs rely on the decomposition of graph Laplacian and operate on the graph spectra. Thus, they are dependent on the structures of the whole graphs and are not localized in space.
As a result, it is difficult to apply them to graphs with varying characteristics or large-scale graphs. On the other hand, spatial-based GNNs define each node's receptive field based on the node's spatial relations and implement localized graph convolution within this field. Therefore, they can overcome the aforementioned limitations and have achieved impressive results on various graph learning tasks, ranging from node classification \cite{hamilton2017inductive}, link prediction \cite{schutt2017schnet,zhang2018link}, to graph classification \cite{dai2016discriminative}.

Though various spatial GNNs have been proposed, such as diffusion GNN  \cite{velivckovic2017graph,atwood2015diffusion, dong2023rethinking, zhang2021nested}, most GNNs follow the neighborhood aggregation scheme which updates node representations by aggregating transformed
representations of nodes and their one-hop neighboring nodes. However, as discussed in the work of Xinyi and Chen \cite{xinyi2018capsule}, a major limitation of current aggregation methods is that they are not expressive enough to capture the presence of different graph sub-structures while at the same time encoding their specific properties, such as connections and positions.  For instance, Kearnes et al. \cite{kearnes2016molecular}, Kipf and Welling \cite{kipf2016semi}, Zhang et al. \cite{zhang2018end} take the mean of transformed node features as the aggregation function, yet Xu et al. \cite{xu2018powerful} has shown that though the mean aggregator can learn the distribution of node representations from the receptive field, it fails to distinguish some simple structures. As such, more expressive aggregators are required to extract useful features characterizing the rich information encoded in graphs. 

While the neighborhood aggregation effectively learns node representations, another challenge in graph classification is to learn graph-level representation from node representations, and various graph pooling modules\cite{zhang2018end,dong2023interpreting,ying2018hierarchical} are proposed for this purpose. In general, pooling modules fall into three categories: summing-based approaches, sorting-based approaches, and hierarchical approaches. 

Summing-based pooling modules \cite{henaff2015deep} generate graph representations by summing nodes' representations extracted through graph convolution operations. Sorting-based pooling modules convert non-Euclidean graph data into grid data by sequentially reading nodes in a meaningful and consistent order. For instance,  PATCHY-SAN \cite{niepert2016learning} generates node order via a node labeling procedure and a canonicalization process to select fixed-sized patches with a fixed number of ordered neighbors for each node.
DGCNN \cite{zhang2018end} proposes to sort nodes for pooling according to their structural roles within the graph. However, since the summing-based graph pooling and the sorting-based graph pooling stack multiple graph convolution layers to propagate information and globally implement the graph down-sampling via a pooling module, the generated graph representation is inherently flat. In order to extract hierarchical graph representations, DiffPool \cite{ying2018hierarchical} uses different GNNs to separately implement neighborhood aggregation and graph pooling, and it provides a framework to hierarchically pool nodes across a broad set of graphs. 

To address the aforementioned limitations and challenges, we propose Subgraph Pattern Graph Neural Network (SPGNN), an end-to-end deep learning architecture for graph learning. SPGNN recognizes salient subgraph patterns from graphs
with an injective graph convolution layer as well as a novel layer-wise graph pooling module. More specifically, we make the following contributions in this paper.
\begin{itemize}
\item For the convolution layer in SPGNN, we propose a novel graph convolution operation that generates high-quality node embeddings to preserve valuable graph topology and node information by injectively updating node representations based on the set of transformed neighboring nodes' representations. 

\item In SPGNN, we invent a novel graph pooling module, WL-SortPool, to generate representations of informative subgraphs with different sizes. WL-SortPool separately arranges unsorted node representations from different graph convolution layers/iterations into a consistent order, bridging the gap between graph data and Euclidean data. This pooling module enables
subsequent conventional neural network layers to capture important subgraph patterns 
from the extracted representation. 

\item We empirically validate the effectiveness of SPGNN on benchmark graph classification datasets. Our results show that the proposed architecture achieves state-of-art results compared to graph kernel approaches and other GNN models for graph classification. 
\end{itemize}

%% file: 2_related.tex
GNNs have become one of the most common classes of approaches for learning tasks over graphs. There are two main components in a GNN model. 1) \textbf{Graph convolution layers} iteratively update node representation from the features of the nodes and their neighboring nodes by differentiable message passing functions, such as recursive updating functions \cite{cho2014learning}, graph attention functions  \cite{velivckovic2017graph}, and loopy belief propagation functions  \cite{dai2016discriminative}. Stacking multiple message passing layers can effectively generate node embeddings that encode localized sub-structures. 2) \textbf{Graph pooling modules:} Gao and Ji \cite{gao2019graph}, Lee et al. \cite{lee2019self}, Zhang et al. \cite{zhang2018end} are interleaved with message passing layers to aggregate graph representation from the extracted node representations for graph-level learning tasks. Due to the expressive power of GNNs to capture graph structures, GNNs outperform other graph heuristics, such as graph kernels  \cite{borgwardt2005shortest,shervashidze2011weisfeiler}, and have achieved state-of-art performance on node classification, graph classification tasks, and link prediction tasks.                   


Another related previous work is the Weisfeiler-Lehman (WL) subtree kernel  \cite{shervashidze2009efficient}, a powerful local pattern induced graph kernel for graph classification. WL subtree kernel leverages the well-known WL algorithm as a subroutine to extract localized subtree patterns from graphs, and it efficiently measures the similarities between graphs. Spatial GNNs show a close relationship with the WL test, and Xu et al. \cite{xu2018powerful} has proven that pooling-assisted GNNs are at most as powerful as the WL test for distinguishing graph structures. However, a major limitation of the WL subtree kernel is its ignorance of the dependency between substructures, and various deep graph kernels have been proposed to address this limitation. For instance, Yanardag and Vishwanathan \cite{yanardag2015deep} proposed a deep graph kernel using models from natural language processing (skip-gram or continuous bag-of-word (CBOW)) to learn latent representations that capture the substructure similarity. Kondor and Pan \cite{kondor2016multiscale} proposed a multi-scale Laplacian graph kernel to capture different levels of substructure similarity and measure the similarity between graphs more globally. However, only limited works have studied this challenge in an end-to-end learning framework.

%% file: 3_settings.tex
\begin{figure}[t]
\centering
\includegraphics[width=0.99\linewidth]{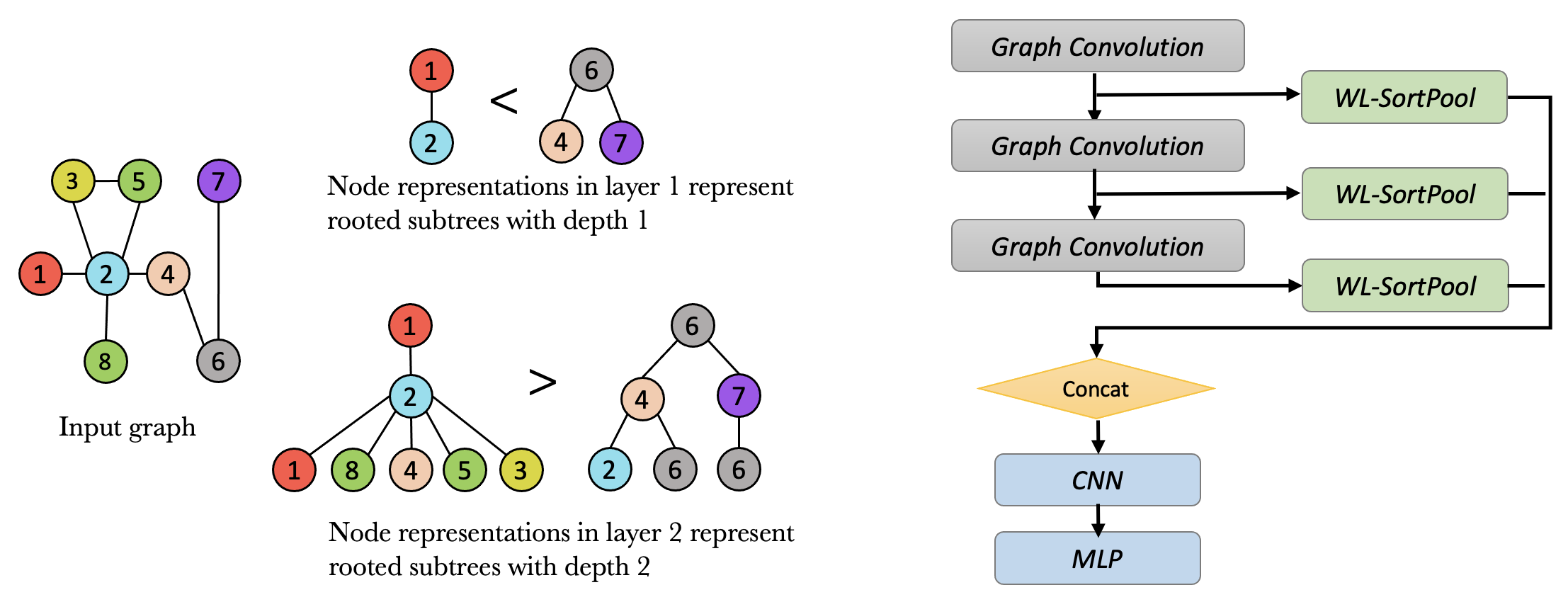}
\caption{ Motivation for the layer-wise sort-pooling module (left) and architecture of SPGNN (right). In the left figure, the input vertices are illustrated as colored nodes. Suppose that the relative importance of rooted subtrees in the same layer is decided by their sizes, node 6 is more important than node 1 in the first layer, while node 1 plays a more important structural role in the second layer. The right figure shows key components of SPGNN that are explained below.}
\label{fig:architect} 
\end{figure}

In this work, we study the classification problem for the undirected unweighted graphs whose adjacency matrix $A$ is a symmetric $0/1$ matrix. In general, the graph classification task takes as inputs a set of graphs $\{G_{1}, G_{2}, ... G_{N}\}$ as well as their classes $\{y_{1},y_{2},...y_{N}\}$, and we aim to develop neural networks that take these graphs as inputs to learn the graph embedding based on which the class of the input graph is predicted. In this paper, we focus on spatial-based neural networks, which directly designs the graph convolution based on spatial relations among nodes, and is thus straightforward to be generalized to classification tasks with directed graphs.

The overall architecture of Subgraph Pattern Graph Neural Network (SPGNN) is illustrated on the right side of Figure \ref{fig:architect}. We explain its details below.

\subsection{Concatenation based injective graph convolution}
\label{sec: deep set graph conv}
In this work, we focus on the most common spatial graph convolution framework, namely the neighborhood aggregation framework, which iteratively updates the representation of a node by propagating information from the node’s neighborhood. Formally, this procedure can be written as:
\begin{equation}
Z^{(l+1)} = f(t(A)Z^{(l)}W^{(l)})
\end{equation}
where $Z^{(l)} \in \mathbb{R}^{n \times d_{l}}$ denotes the node representation matrix in layer $l$, and $W^{(l)} \in \mathbb{R}^{d_{l} \times d_{l+1}}$ denotes the trainable parameter matrix of feature transformation. Each row $i$ of $Z^{l}$ is the representation of node $i$ in layer $l$. $t(A)$ is the neighborhood aggregator built on the adjacency matrix $A$, which defines the information aggregation rule. In previous works, a common choice of $t(A)$ is the normalized adjacency matrix (i.e. $t(A) = \tilde{D}^{-\frac{1}{2}}\tilde{A}\tilde{D}^{-\frac{1}{2}}$ where $\tilde{A}=A+I$ is the adjacency matrix with self-loop and $\tilde{D}$ is the corresponding degree matrix) and its variants. The resulting aggregator will take the average over transformed node features in local neighborhoods.\\

\textbf{GNNs and Weisfeiler-Lehman test.} 
The neighborhood aggregation strategy has a close connection to the Weisfeiler-Lehman (WL) graph isomorphism test \cite{shervashidze2009efficient}, an efficient tool for graph isomorphism checking. The WL algorithm iteratively updates a node's label/color by compressing its 1-hop neighbors’ labels, lexicographically sorting the compressed labels, and inductively mapping sorted labels into new short colors. This process is repeated until the multiset of node colors for two graphs are different, which indicates two graphs are not isomorphic, or until the number of iteration reaches the number of nodes in these two graphs. Similarly, the above neighborhood aggregation process can be interpreted in some way analogous to WL. The feature transformation $Y^{(l)}=Z^{(l)}W^{(l)}$ plays a similar role to the label compression step and the label sorting step in the WL test, while the nonlinear transformation of aggregated features $f(t(A)Y^{l})$ can be viewed as the new label generation step. In fact, \cite{zhang2018end} has shown that we can treat graph convolution in (1) as a 'soft' version of the WL test, and the generated node representations $Z^{(l+1)}$ can be interpreted as continuous WL colors. Hence, node representations (i.e. continuous WL colors) from different graph convolution layers represent rooted subtrees with different depths.

A significant limitation is that the sum/mean based aggregator fails to distinguish the central node from the neighboring nodes. Therefore, this simple heuristic is not expressive enough to provide an injective aggregation scheme that maps different node neighborhoods to different updated representations. For instance, let $x_{a},x_{b},x_{c}$ be transformed node representations such that $x_{a} \neq x_{b} = x_{c}$, and $G_{1}=(V_{1},E_{1})$, $G_{2}=(V_{2},E_{2})$ be two subgraphs such that $V_{1}=V_{2}=\{a,b,c\}, E_{1}=\{(a,b),(a,c)\}, and \ E_{2} = \{(b,a),(b,c)\}$. Then the sum based aggregator will generate the same updated representation for distinct rooted subtrees $G_{1}$ and $G_{2}$). To address this challenge, \cite{xu2018powerful} develops an expressive architecture, Graph Isomorphism Network (GIN), to maximize the discriminative power among GNNs by an injective neighborhood aggregation function:
\begin{equation}
\label{equ: gin} 
Z^{(l+1)}_{i,:} = \textit{MLP}^{(l)}((1+\epsilon^{(l)})Z^{(l)}_{i,:}  + \sum_{j \in \mathcal{N}(i)}Z^{(l)}_{j,:})
\end{equation}
Where $\epsilon$ is a trainable parameter. However, although GIN is provably injective for infinitely many choices of $\epsilon$, it could be practically difficult to identify the optimal $\epsilon$ in a deep learning framework. According to the reported experimental results in the original paper, GIN-0 (i.e. $\epsilon = 0$) outperforms GIN-$\epsilon$ (i.e. $\epsilon$ is a trainable parameter) on all the testing datasets
, yet the neighborhood aggregation function (2) is not injective when $\epsilon$ is $0$ since the MLP takes the summed node representations in the neighborhood as inputs. Hence, it is desirable to develop the injective aggregation function without parameters empirically hard to optimize. \\


\textbf{Concatenation based graph convolution.} Here we present our proposed graph convolution scheme. Inspired by the graph attention mechanism  \cite{velivckovic2017graph} which utilizes concatenation to bring two features (the transformed representations of a central node and that of a neighboring node) together while keeping their identifications, we propose a concatenation-based aggregator (Cat-Agg) which operates on sets of features and show that Cat-Agg results in injective graph convolutions. 

Let $S = \{S_{n} | n \in 1,2,...N\}$ be a set of d-dimensional features, and $S_{i} \in S$ be a pre-selected element in the set $S$. Cat-Agg takes the following form:
\begin{equation}
\textit{Cat-Agg}(S_{i}, S) = \sigma( M^{T} \it{concat}\left(S_{i}, \it{sumpool}(\{S_{j} | j \neq i \}) \right) \ )
\end{equation}
where $M \in \mathbb{R}^{2 \times d, d_{out}}$ denotes the trainable parameter matrix that provides sufficient expressive power to extract aggregated features, and $\sigma$ denotes a nonlinear activation function.

In the proposed aggregator, \textit{sumpool} is used to extract the representative feature of the subset $\{S_{j} | j \neq i \}$.  \textit{concat} concatenates $S_{i}$ and the pooled feature. Cat-Agg blends information from the selected element $S_{i}$ and that from the rest of the elements in $S$. In order to obtain sufficient expressive power to generate a high-level aggregated feature, at least one linear layer is applied after the concatenation operation. As Proposition \ref{prop: inject} will show, it is sufficient to generate the injective graph convolution when we apply one-layer Cat-Agg in the neighborhood aggregation framework. Hence, we only take one linear layer to reduce the number of parameters and avoid overfitting.




\begin{figure*}[th]
\begin{center}
\centerline{\includegraphics[width=0.75\textwidth]{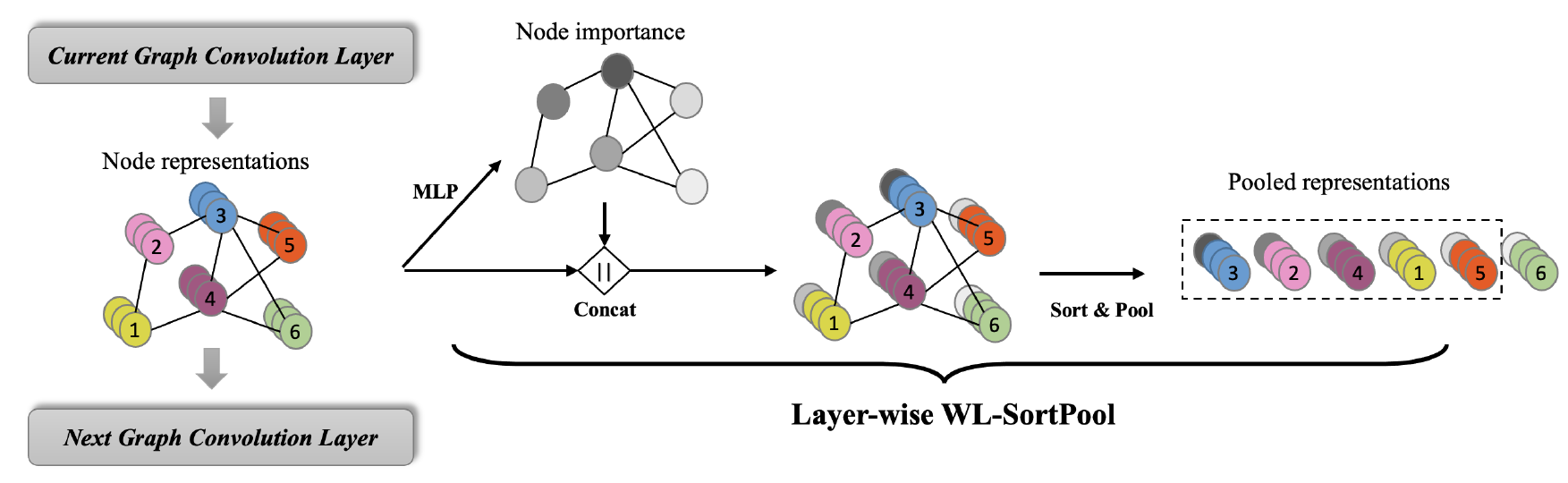}}
\caption{ Illustration of WL-SortPool: In each layer/iteration, the proposed graph convolution updates the node representations.
After that, WL-SortPool learns node importance. Node representations in the current layer are sorted and pooled accordingly.}
\label{fig: wl sortpool}
\end{center}
\end{figure*}

With the Cat-Agg aggregator to read out information from a set of features, we can apply it in the neighborhood aggregation framework to update node representations by aggregating transformed node representations from their neighborhoods. The resulting graph convolution is constructed as following: 
\begin{equation}
Z_{i,:}^{(l+1)} = \textit{Cat-Agg}(Y_{i,:}^{(l)}, \tilde{\mathcal{N}}(i))
\end{equation}
where $Y^{(l)}=\sigma (Z^{(l)}W^{(l)}) \in \mathbb{R}^{n \times d_{l+1}}$ denotes a nonlinear activation of the transformed node representation matrix. Let $\mathcal{N}(i)$ be the set of neighboring nodes of the node $i$, then $\tilde{\mathcal{N}}(i) = \{Y_{i,:}^{(l)}\} \cup \{ Y_{j,:}^{(l)} | j \in \mathcal{N}(i) \}$. We can using other pooling methods, such as the max-pool and mean-pool, to get other variants, as long as the alternative pooling method is invariant to the order to read elements in the set and the resulting operation over these elements is communicative. Since Cat-Agg does not depend on the order to read set elements, the proposed graph convolution in (4) is invariant to the permutation of the order of nodes in the input graph. 

\begin{prop}
Assume Cat-Agg has one linear layer, the corresponding graph convolution in (4) can approximate the neighborhood aggregation function in (2) for any fixed scalar $\epsilon$.
\label{prop: inject}
\end{prop}

Proposition \ref{prop: inject} is straightforward due to the the universal approximation theorem \cite{hornik1989multilayer,hornik1991approximation}. As such, the proposed graph convolution can satisfy the injectiveness condition.


\subsection{The WL-SortPool mechanism}

Given node representations from all graph convolution (GC) layers/iterations, the node-level tasks such as node classification and link prediction are implementable by directly feeding node representations from the last GC layer into any loss function. However, for graph-level tasks like graph classification, another major challenge is to extract expressive graph representation from these node representations. 



In an earlier work, SortPooling \cite{zhang2018end} indicates that sorting nodes in a consistent order to read help extract global information from graphs. SortPooling reads nodes in a meaningful order by concatenating nodes representations from all GC layers and lexicographically sorting the last GC layer's output.
SortPooling uses a single mechanism to generate the same order for node representations from different GC layers. However, as Figure \ref{fig:architect} shows, since different layers' outputs capture nodes' local structures of different scales, they may have different relative importance relation for the graph classification purpose, 
Hence, a unified pooling strategy across all layers may introduce a loss of structural information, and sorting them separately could enable us to obtain the global graph topology information based on different granularities of local structures. 

To reduce the loss of structural information, we propose a novel pooling module called WL-SortPool which layer-wise sorts nodes representations (WL continuous colors), as illustrated in Figure \ref{fig: wl sortpool}. Inspired by DGCNN, we assume that node representations in continuous WL colors inherently encode the relative structural importance of the corresponding rooted subtrees. We propose to learn such importance through a multiple layer perceptron (MLP), which takes as input the node representations generated by the GC layers. 

In each layer $l$ of SPGNN, the node representation matrix $Z^{(l)} \in \mathbb{R}^{n \times d_{l}}$ is generated through the proposed graph convolution. To obtain the node order in the current layer, $Z^{(l)}$ is fed into a MLP to learn the importance of nodes (i.e. $I^{(l)} = MLP(Z^{(l)})$), and this computed importance vector is then concatenated with the node representation matrix: $H^{(l)} = \textit{concat}(Z^{(l)},I^{(l)})\in \mathbb{R}^{n \times (d_{l} + 1)}$). Hence, the last feature dimension in $H^{(l)}$ encodes the relative structural importance between nodes in layer $l$. As such, WL-SortPool layer-wise sorts $H^{(l)}[:,-1]$ in a descending order and keeps the $k$ most important node representations (henceforth this operation is denoted by $\textit{sort-k}(H^{(l)}) \in \mathbb{R}^{k \times (d_{l}+1)}$) by truncating the $n-k$ least important nodes or extending $k-n$ nodes with zero embeddings. After the layer-wise pooling, the graph representation is achieved by concatenating $H^{(l)}$ across all layers: 
\begin{equation}
Z_{G} = \textit{concat}\left(\textit{sort-k}(H^{(l)}) \ |\ l = 1,2,...L  \right) 
\end{equation}

As such, WL-SortPool separately generates the order of continuous WL colors
in different layers via learning the node importance based on node representations from the current GC layer. As a result, it leads to distinct mechanisms to generate node orders in different layers. Hence, SPGNN is more effective in maintaining global graph topology information 
and recognizing salient subgraph patterns in graphs. Besides the injectiveness condition, another important criterion for graph representation learning is that the learning framework should be invariant to varying node orderings of the input graph, i.e. isomorphic graphs should be mapped to the same graph representation. Theorem \ref{theorem: dsgnn} indicates that WL-SortPool is invariant to the permutation of input node ordering and satisfies the permutation invariance condition.   

\begin{theorem}
For any two graphs $G_{1}$ and $G_{2}$ that are  isomorphic, their graph representations generated by the proposed graph convolution and WL-SortPool are identical, i.e. $Z_{G_{1}} = Z_{G_{2}}$.
\label{theorem: dsgnn}
\end{theorem}

\noindent $\it{Proof:}$ As section \ref{sec: deep set graph conv} shows, the proposed graph convolution is invariant to permutation of node orderings. Thus, there exists a permutation matrix $P_{\pi}$ such that $P_{\pi}H^{{(l)}}_{G_{1}} = H^{{(l)}}_{G_{2}}\ for\ \forall l $. In each layer $l$, the node sorting operation can guarantee that a node pair from the two multisets $\{\textit{rows of} \ H^{{(l)}}_{G_{1}} \}$ and $\{\textit{rows of} \  H^{{(l)}}_{G_{2}}\}$  have a tie in the pooled representation if and only if they have the same node importance, i.e. the same node representation. Therefore, these two graphs will have the same pooled representation in each layer (i.e. $\textit{sort-k}(H^{{(l)}}_{G_{1}}) = \textit{sort-k}(H^{{(l)}}_{G_{2}}) \ \ \forall l$). By concatenating pooled representations from different layers, the resulting graph representations are the same for $G_{1}$ and $G_{2}$.



\subsection{Comparison to other algorithms}
The proposed SPGNN architecture directly addresses the neighborhood aggregation challenge and the graph pooling challenge for graph classification. Here we  give some comparison between SPGNN
with some related algorithms. 

\textbf{Comparison with GIN.} Both SPGNN and Graph Isomorphism Network (GIN)  \cite{xu2018powerful} propose an injective neighborhood aggregation scheme. However, the aggregation function (2) in GIN introduces a trainable parameter $\epsilon$ to satisfy the injectiveness condition, which makes the model difficult to train and empirically reduces the performance. Compared to GIN, SPGNN utilizes a concatenation operation in the aggregation function (4) to distinguish the central node from the neighboring nodes. As \ref{sec: deep set graph conv} indicates, the concatenation operation can help generate the injective graph convolution, while introducing no extra training difficulty in the deep model. 

\textbf{Comparison with SAGpool.} The key innovation of WL-SortPool is to layer-wise sort node representations according to their structural roles. A previous work, SAGPool  \cite{lee2019self}, also introduces a layer-wise graph pooling module. SAGPool uses GNNs to learn the projection and self-attention scores to perform graph pooling which selects the top-ranked nodes and filters out the rest. SAGPool alternately stacks graph convolution layers and pooling layers. As such, it iteratively operates on coarser and coarser graphs. In comparison, as SPGNN separately implements graph convolution and graph pooling, its WL-SortPool layer only reads out graph representations from node representations and will not affect the graph convolution process.

%% file: 4_evaluation.tex
\begin{table*}[t]
\caption{Comparison with Graph Kernel Methods}
\label{table graph kernel}
\centering
\resizebox{\textwidth}{!}{
\begin{tabular}{lcccccccc}
\toprule
Dataset & MUTAG & PTC & NCI1 & PROTEINS & D$\&$D & IMDB-B & IMDB-M & RE-M5K\\
\midrule
\textbf{SPGNN}  & \textbf{84.39}$\pm$ \textbf{0.19}& \textbf{62.29}$\pm$ \textbf{0.75} & 76.37 $\pm$0.45 & \textbf{76.00} $\pm$ \textbf{0.53} & \textbf{79.16} $\pm$ \textbf{0.58} & \textbf{72.77} $\pm$ \textbf{0.32} & \textbf{50.45} $\pm$ \textbf{0.55} & \textbf{50.95} $\pm$ \textbf{0.59}\\
\midrule
GK & 81.39$\pm$ 1.74& 55.65$\pm$ 0.46& 62.49$\pm$ 0.27 & 71.39$\pm$ 0.31 & 74.38$\pm$ 0.69 & 65.87 $\pm$ 0.98 & 43.89 $\pm$ 0.38 &41.01  $\pm$ 0.17\\
RW    & 79.17$\pm$ 2.07& 55.91$\pm$ 0.32& $>$ 3 days & 59.57$\pm$ 0.09 & $>$ 3 days & $-$ & $-$ & $-$\\
PK    & 76.00$\pm$ 2.69& 59.50$\pm$ 2.44& 82.54$\pm$ 0.47 & 73.68$\pm$ 0.68 & 78.25$\pm$ 0.51 & $-$ & $-$ &\\
WL   & 84.11$\pm$ 1.91& 57.97$\pm$ 2.49& \textbf{84.46}$\pm$ \textbf{0.45} & 74.68$\pm$ 0.49 & 78.34$\pm$ 0.62 & \textbf{72.86} $\pm$ \textbf{0.76} & \textbf{50.55} $\pm$ \textbf{0.55}& 49.44  $\pm$ 2.36\\
\bottomrule
\end{tabular}
}
\end{table*}

\begin{table*}[t]
\caption{Comparison with Deep Learning Methods}
\label{table deep learning}
\centering
\resizebox{\textwidth}{!}{
\begin{tabular}{lccccccccc}
\toprule
Dataset  & PTC & NCI1 & PROTEINS & D$\&$D & IMDB-B & IMDB-M & COLLAB & RE-M5K\\
\midrule
\textbf{SPGNN}    & 62.29 $\pm$ 0.75 & \textbf{76.37} $\pm$ \textbf{0.45} & \textbf{76.00} $\pm$ \textbf{0.53} &  79.16 $\pm$ 0.58  & \textbf{72.77} $\pm$ \textbf{0.32} & \textbf{50.45} $\pm$ \textbf{0.55} & 77.08 $\pm$ 0.59 & \textbf{50.95} $\pm$ \textbf{0.59} \\
\midrule
DGK  & 60.08 $\pm$ 6.55  & 62.48$\pm$ 0.25& 71.68$\pm$ 0.50& 73.50 $\pm$ 1.01 &66.96$\pm$ 0.56 &44.55$\pm$ 0.52 &73.09 $\pm$ 1.42 &41.27  $\pm$ 0.78 \\
PSCN  & 62.29 $\pm$ 5.68  & 76.34$\pm$ 1.68&  75.00 $\pm$ 2.51& 76.27$\pm$ 2.64 &71.00 $\pm$ 2.29 &45.23 $\pm$ 2.84 &72.61 $\pm$ 2.51 &49.10  $\pm$ 0.70\\
Graph2Vec & 60.17 $\pm$ 6.86 & 73.22 $\pm$ 1.81 & 73.30 $\pm$ 2.05 & 71.10 $\pm$ 4.02 & $-$ & $-$ & $-$ & $-$\\
\midrule
DCNN(sumPool) & 56.60$\pm$ 2.89 & 56.61$\pm$ 1.04&  61.29 $\pm$ 1.60& 58.09 $\pm$ 0.53 &49.06$\pm$1.37 &33.49 $\pm$ 1.42 &52.11 $\pm$ 0.71 & $-$ \\
GIN(sumPool)  & \textbf{64.1} $\pm$ \textbf{7.51} & 76.25  $\pm$ 2.63 &  75.56  $\pm$ 3.62 & \textbf{80.16} $\pm$ \textbf{1.91} & \textbf{72.69} $\pm$ \textbf{4.1} & 50.01  $\pm$ 4.42 & 73.50 $\pm$ 1.49 & \textbf{50.98} $\pm$ \textbf{2.60} \\ 
GAT(sumPool)   & 58.52$\pm$ 1.69 & 70.61$\pm$ 0.88 &  73.71$\pm$ 1.16 & 74.38$\pm$ 1.52 & 70.65$\pm$ 1.17 & 48.91$\pm$ 1.01 & 69.17$\pm$ 1.47 & 41.15$\pm$ 0.96\\ 
SAGPool & $-$ & 74.18  $\pm$ 1.20 &  71.86  $\pm$ 0.97 & 76.45  $\pm$ 0.97 & $-$ & $-$ & $-$ & $-$\\
gPool & $-$ & 70.02 $\pm$ 2.21 & 71.10 $\pm$ 0.91 & 75.01 $\pm$ 0.86 & $-$ & $-$ & \textbf{77.36} $\pm$ \textbf{1.56} & 50.02 $\pm$ 0.89 \\
DGCNN(sortPool) & 58.59$\pm$ 2.47 & 74.44$\pm$ 0.47&  75.54$\pm$ 0.94& 79.37 $\pm$ 0.94 &70.03$\pm$ 0.86 & 47.83 $\pm$ 0.25 &73.96 $\pm$ 0.49 & 48.70  $\pm$ 4.54\\
\bottomrule
\end{tabular}
}
\end{table*}

\begin{table*}[t]
\caption{Ablation Study}
\label{ablation}
\centering
\resizebox{\textwidth}{12mm}{
\begin{tabular}{lcccccccc}
\toprule
Architecture configuration  & PTC & PROTEINS & D$\&$D & IMDB-B & IMDB-M & COLLAB & Ave. Accuracy\\
\midrule
SPGNN   & \textbf{62.29} $\pm$ \textbf{0.75} & \textbf{76.00} $\pm$ \textbf{0.53} &  79.16 $\pm$ 0.58  & \textbf{72.77} $\pm$ \textbf{0.3} & \textbf{50.45} $\pm$ \textbf{0.55} & \textbf{77.08} $\pm$ \textbf{0.59} & \textbf{69.63} \\
\midrule
Architecture 1: proposed graph conv + sortpool & 55.88 $\pm$ 0.01 &  74.26 $\pm$ 0.24 & \textbf{79.35}$\pm$ \textbf{0.51} &72.25 $\pm$ 0.35 & \textbf{50.81} $\pm$ \textbf{0.15} & 76.45 $\pm$ 0.25 & 68.17\\
Architecture 2: GIN graph conv + sortpool  & 55.88 $\pm$ 0.01 &  73.15 $\pm$ 0.58 & 79.01 $\pm$ 0.24 & 71.45 $\pm$ 0.16 & 49.39 $\pm$ 0.27 & 76.24 $\pm$ 0.26 & 67.52\\
DGCNN   & 58.89$\pm$ 2.47&  75.04$\pm$ 0.88& 79.00 $\pm$ 0.63 &70.03$\pm$ 0.86 & 47.83 $\pm$ 0.25 &73.96 $\pm$ 0.49 & 67.46 \\
\bottomrule
\end{tabular}
}
\end{table*}

We evaluate SPGNN on multiple benchmark datasets for graph classification by comparing its performance against  the state-of-art graph kernel methods and other deep learning approaches.  

\subsection{Datasets}
We choose five biological informatics benchmark datasets: MUTAG, PTC, NCI1, PROTEINS, D$\&$D, and four social network benchmark datasets: COLLAB, IMDB-BINARY (IMDB-B), IMDB-MULTI (IMDB-M), REDDIT-MULTI5K (RE-M5K). All the benchmark datasets contain undirected graphs. As for node features, biological datasets have input node labels, whereas social network datasets have uninformative node features, i.e. all node labels are set to be 0. Following previous works  \cite{zhang2018end,xu2018powerful,verma2018graph}, we take node degrees as node labels when nodes in the input graph do not have labels. 
  
\subsection{Model configuration}
In our experiments, we use a single network structure for all datasets to make a fair comparison with graph kernel baselines and other deep-learning based methods. For all configurations, the proposed SPGNN uses 4 graph convolution layers with 32 output feature channels, the sum-pool-based Cat-Agg for aggregation, and a 2-layer MLP with $16$ hidden units in WL-SortPool to learn the node importance. 

Similar to DGCNN, we provide a simple scheme to select $k$ in WL-SortPool. (1) For datasets with small graphs (the average number of nodes $<$ 30), $k$ is set to $30$. (2) For datasets with large graphs (the average number of nodes $> 200$), we set $k$ such that $50\%$ graphs in biological datasets or $90\%$ graphs in biological datasets have nodes less than $k$. We keep more nodes in social network datasets to reduce the structural information loss since no input node features are available and the graph classification task mainly relies on the graph structure. (3) For the other datasets with an average node number between 30 and 200, we search $k$ in $\{30, 50\}$. Furthermore, we apply L2 regularization with $\lambda=0.05$ for datasets with relatively large graphs as defined in case (2).

In SPGNN, the standard CNN part consists of two 1-D convolutional layers and one 1-D MaxPooling layer. The two convolutional layers respectively have the following hyper-parameter settings: the output channels are $(16,32)$, the filter sizes are $(32 \times 4, 5)$, the step sizes are $(32 \times 4, 1)$, and the MaxPooling layer has a filter size of $2$ and a step size of $2$. Finally, this CNN architecture is followed by an MLP with one hidden layer with $100$ hidden units and a dropout layer with a $0.5$ dropout rate. The number of training epochs is searched in $\{50,100,150,200\}$ by selecting the one with the maximum average validation accuracy. 

To provide robust model performance, we perform 10-fold cross-validation with LIB-SVM \cite{chang2011libsvm}, and report the accuracy averaged over 10 folds and the standard deviation of validation accuracies across the 10 folds.

\subsection{Baseline methods}

\textbf{Graph kernel methods:} Graph kernels measure the graph similarity by comparing substructures in graphs that are computable in polynomial runtime. Informally, graph kernel algorithms exploit the graph topology by decomposing graphs into subgraphs and designing corresponding kernels to build the graph embedding.  We compare our deep learning architecture with following four graph kernels: Random Walk graph kernel (RW) \cite{vishwanathan2010graph}, Graphlet Kernel (GK) \cite{shervashidze2009efficient}, Shortest Path graph kernel (PK) \cite{borgwardt2005shortest}, and Weisfeiler-Lehman Sub-tree Kernel(WL) \cite{shervashidze2011weisfeiler}. By default, we report the testing results provided in \cite{zhang2018end}, where the conventional setting is adopted as in previous works. For WL, the height parameter is selected from the set $\{$1, 2, \ldots, 5$\}$. For PK, the height parameter is searched in the same set and set the bin width to be $0.001$. For GK, the graphlet size is set to be $3$. For RW kernel, the decay parameter $\lambda$ is selected from $\{10^{-1}, 10^{-2}, \ldots, 10^{-5} \}$. Furthermore, 10-fold cross validation is performed and we report the average accuracy and the standard deviations.  
 
\textbf{Deep learning methods.} We also compare SPGNN with nine deep-learning based algorithms, which can be categorized into two types: 1) two-phases deep learning algorithms that require additional preprocessing steps before implementing the classification tasks ( PATCHY-SAN (PSCN) \cite{niepert2016learning}, Graph2Vec \cite{narayanan2017graph2vec}, Deep Graph Kernel (DGK) \cite{yanardag2015deep}); 2) end-to-end GNNs that take the basic graph convolution framework to aggregate nodes' information and then implement a global graph pooling to readout (Diffusion-CNN (DCNN) \cite{atwood2015diffusion}, Graph Attention Network (GAT)  \cite{velivckovic2017graph},Deep Graph CNN (DGCNN) \cite{zhang2018end}, Graph Isomorphism Network (GIN)  \cite{xu2018powerful}); 3) end-to-end GNNs based on hierarchical graph pooling techniques (Self-Attention Graph Pooling (SAGPool) \cite{lee2019self}, Graph U-Nets (gPool) \cite{gao2019graph}).

Except for GIN and GAT, we report testing results provided in  \cite{zhang2018end,xinyi2018capsule,lee2019self}. GAT is originally tested for the node classification problem, and no graph classification results are reported. In our experiments, we stack 4 graph attention convolution layers with 32 output feature channels and concatenate sum-pooled features from the layers to generate the graph representation, which is then passed to an MLP to predict the graph label. GIN does not use the validation set to report the testing results in the original work. Instead, GIN computes the average of validation curves across 10 folds, and then selects the epoch that maximizes averaged validation accuracy. For a fair comparison, we do not take such a testing scheme. Instead, we search for the optimal number of training epochs in $\{50, 100, 150, 200, 250, 300, 350\}$. Other hyperparameters are tuned following the original setting: the number of hidden units is $32$ for bioinformatics graphs and $64$ for social graphs, the batch size is $50$, and the dropout rate is $0.5$.


\subsection{Experimental results}

Table~\ref{table graph kernel} and Table~\ref{table deep learning} list the experimental results, and the best-performing methods are highlighted in bold. More specifically, if the difference of the average classification accuracy between the best and the second-best methods is smaller than $0.2 \%$, and the t-test at significance level $5\%$ cannot distinguish them, we think these two methods work equivalently well and highlight both of them.


\textbf{Comparison with graph kernels.} As Table\ref{table graph kernel} shows, SPGNN outperforms all graph kernels on five datasets: MUTAG, PTC, D$\&$D, PROTEINS, RE-M5K, and obtains highly competitive results with the state-of-art graph kernel, WL kernel, on two benchmark datasets: IMDB-B, IMDB-M. As such, though SPGNN is tested with a single structural setting, it is consistently competitive with graph kernel methods that tune structural hyper-parameters for best performance. Such observation indicates that SPGNN effectively generates graph representations that obtain important subgraph patterns with different scales for graph classification. 

\textbf{Comparison with deep learning methods.} Table\ref{table deep learning} compares SPGNN to other deep learning baselines. SPGNN achieves the state-of-art performance on 5 of 8 datasets (NCI1, PROTEINS, IMDB-B, IMDB-M, RE-M5K), and achieves the top 3 best performance on the rest 3 datasets (second best on PTC, COLLAB, and third best on D$\&$D).(1) SPGNN shows a significant classification accuracy improvement over two-phase deep learning baselines (DGC, PSCN, Graph2Vec), as it enables a unified way to embed and read important local structures with different scales in the end-to-end learning framework that does not require any preprocessing step. (2) SPGNN is able to improve the average accuracy over DGCNN by a margin of $2 \%$, which indicates that WL-SortPool is more effective than SortPool in capturing informative subgraph patterns with different scales by layer-wise sorting node representations. (3) Though it's provable that GIN generalizes WL test and WL subtree kernel, SPGNN can achieve the same average classification accuracy with GIN (SPGNN: $68.13 \%$; GIN: $67.91 \%$), while SPGNN has a significantly smaller average prediction variance. Hence, SPGNN empirically maximizes the representational capacity of basic GNN models by effectively detecting important subgraph patterns. (4) We also observe that SPGNN outperforms GNNs based on hierarchical graph pooling techniques (gPool and SAGPool). The layer-wise sorting mechanism of WL-SortPool allows SPGNN to process subgraphs of different scales (rooted subtrees of different depth) in graphs, separately. Hence, it can recognize salient subgraphs of different scales for the classification task.



\textbf{Ablation study.} In the end, we perform extensive ablation studies to substantiate the effectiveness of the proposed graph convolution and WL-sortpool. For a fair comparison, all architectures take the same number of graph convolution layers, and we set the size (K) of the sortpool and that of the layer-wise WL-sortpool to be same. Without loss of generality, we randomly select 3 biological informatics  datasets (PTC, PROTEINS, and D$\&$D) and 3 social network datasets (IMDB-B, IMDB-M, and COLLAB) to perform the ablation study, and take the average prediction accuracy to measure the effectiveness of architectures. Experimental results are provided in Table \ref{ablation}. (1) DGCNN, architecture 1 and architecture 2 have the same graph pooling method, sortpool. However, architecture 1 and architecture 2 have injective graph convolution layers, while DGCNN has non-injective graph convolution layers. Hence, both architecture 1 and 2 achieve better average prediction accuracy and outperform over DGCNN on 4 out of 6 datasets (D$\&$D,IMDB-B,IMDB-M,COLLAB). (2) Although architecture 1 and architecture 2 have the injective graph convolution layer, the graph convolution layer (function \ref{equ: gin}) in architecture 2 is hard to optimize in practice. Hence, the proposed concatenation based graph convolution layer works better than the injective graph convolution function of GIN (function \ref{equ: gin}). (3) Finally, SPGNN and architecture 1 have the same graph convolution layer but different graph pooling methods. Since SPGNN significantly improve the performance over architecture 2, the proposed layer-wise WL-sortpool can capture more salient graph patterns than sortpool.

%% file: 5_conclusion.tex
In this paper, we have presented a novel graph neural network architecture called SPGNN to effectively generate representations of important subgraph patterns for graph classification. SPGNN addresses the main challenges in graph classification and offers several advantages over existing algorithms. First, the proposed graph convolution in SPGNN can discriminate non-isomorphic subgraph patterns by injectively updating the node representations. Furthermore, we have proposed a novel WL-SortPool module which enables SPGNN to focus on salient subgraph patterns with different scales by sorting node representations in a layer-wise fashion. Extensive empirical results have shown that SPGNN can achieve highly competitive performance and outperform other state-of-art methods on many benchmark datasets. Moreover, we have visualized the subgraph patterns that are recognizable by SPGNN.

In our future work, we will explore the potential of incorporating graph attention mechanisms in the injective graph convolution for a more expressive neighborhood aggregation framework. We will also investigate other pooling modules to recognize more informative graph patterns for various graph learning tasks.

%% file: 6_supplementary.tex
\section{Datasets Description}
Here we introduce details of the benchmark datasets we use in the experiment section.

\begin{table*}[h]
\caption{Biological Informatics Datasets}
\label{bio datasets}
\centering
\begin{tabular}{lcccccc}
\toprule
Dataset & Graphs & Nodes (max) & Nodes (avg.) & Edges (avg.) & Node Labels & Classes\\
\midrule
MUTAG & 188 & 28 & 17.93 & 19.79 & 7 & 2 \\
PTC & 344 & 109 & 25.56 & 27.36 & 11 & 2\\
NCI1 & 4110 & 111 & 29.87 & 34.30 & 23 & 2 \\
PROTEINS & 1113 & 620 & 39.06 & 72.81 & 4 & 2 \\
D$\&$D & 1178 & 5748 & 284.31 & 715.65 & 82 & 2 \\
\bottomrule
\end{tabular}
\end{table*}

\begin{table*}[h]
\caption{Social Datasets}
\label{social datasets}
\centering
\begin{tabular}{lcccccc}
\toprule
Dataset & Graphs & Nodes (max) & Nodes (avg.) & Edges (avg.) & Node Labels & Classes\\
\midrule
IMDB-B & 1000 & 136 & 19.77 & 193.06 & $-$ & 2 \\
IMDB-M & 3500 & 89 & 13.00 & 131.87 & $-$ & 3\\
COLLAB & 5000 & 492 & 74.49  & 4914.99 & $-$ & 3 \\
REDDIT-M5K & 4999 & 3783 & 508.5 & 1189.74 & $-$ & 5 \\
\bottomrule
\end{tabular}
\end{table*}

\section{Visualization}

\begin{figure}[h]
  \centering
  \includegraphics[width=0.35\linewidth]{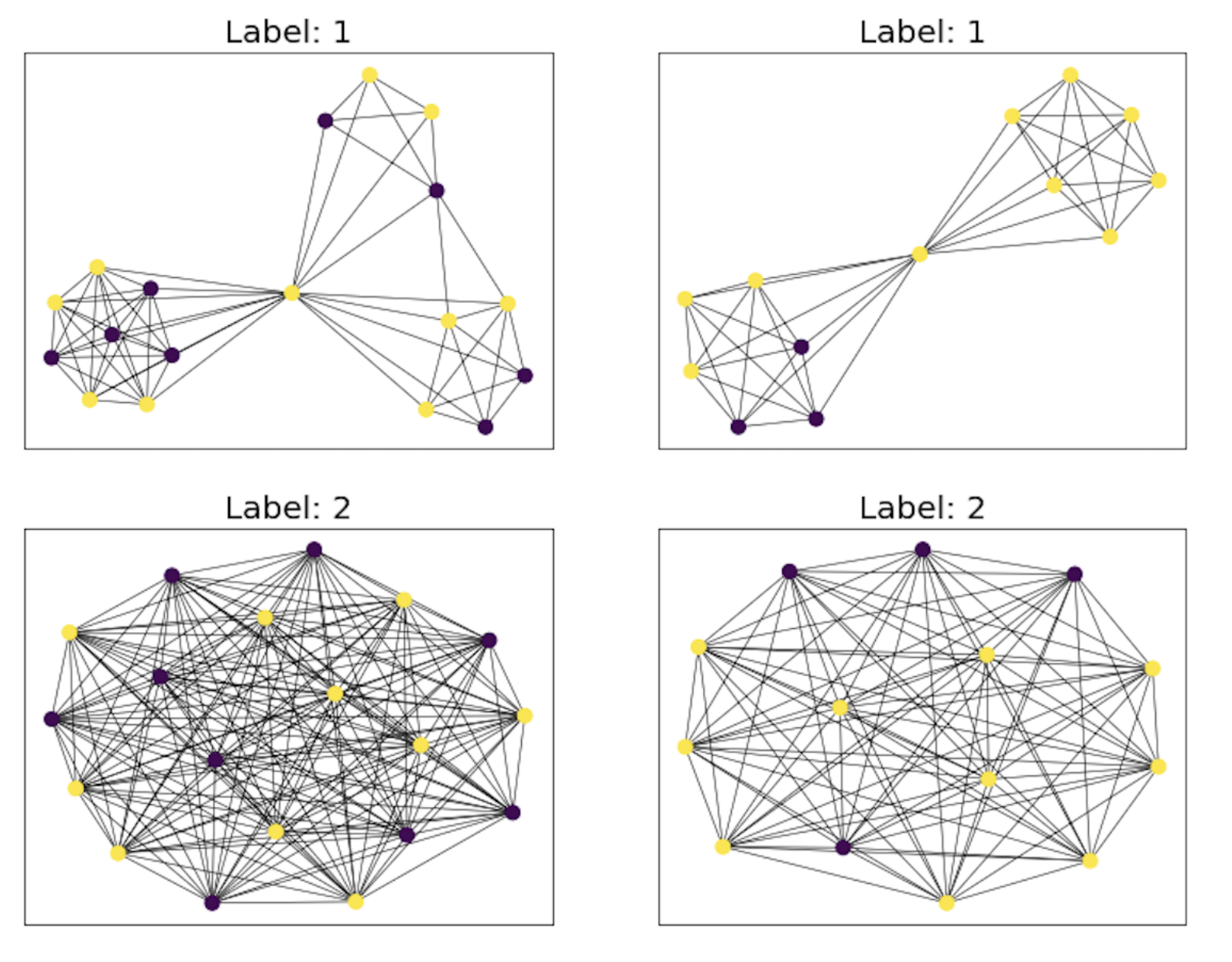}
  \caption{Visualization of layer-wisely pooled nodes based on the IMDB-M dataset.}
  \label{fig: visual}
\end{figure}

We use the relatively small dataset, IMDB-M, to visualize the pooled nodes to see whether SPGNN can obtain salient subgraphs with different scales. In this experiment, SPGNN takes $3$ layers and $k$ is set to $10$. For each node, we count the total number of layers that the node is selected by WL-SortPool, and then rank nodes accordingly and highlight the top 10 nodes in yellow.

As Figure\ref{fig: visual} indicates, WL-SortPool enables the proposed architecture to recognize subgraph patterns with different scales. When the input graphs have community structures  (graphs with label 1 in Figure \ref{fig: visual}), SPGNN not only obtains nodes with high centrality, but also selects representative nodes from communities with different local structural properties. When the input graphs are complete (graphs with label 2 in Figure \ref{fig: visual}), since nodes in IMDB-M have no label, they have equivalent structural roles and receive equal attentions from SPGNN. 

\section{Baselines Description}
In this section, we briedfly describe details of the GNN baselines in the experiment section. 

1) \textbf{Two-phases deep learning algorithms}. PSCN,  Graph2Vec, and DGK require additional preprocessing steps before implementing the classification tasks. In the preprocessing steps, PSCN extracts fixed-size patches for nodes from their neighborhood and needs external software (NAUTY \cite{mckay2014practical}) to generate global node order as well as the total order of nodes within each patch. DGK applies NLP models (continuous bag-of-words and Skip-gram models \cite{mikolov2013efficient}) to generate the latent representation of substructures that model their dependency. Graph2Vec learns graph embeddings in advance in a completely unsupervised manner. After such operations,  standard machine learning algorithms such as kernel-based SVM and standard CNN are applied for graph label prediction.  
  
2) \textbf{GNN baselines.} DCNN, DGCNN, GAT, GIN, and SAGPool follow the basic graph convolution framework that aggregates nodes' information in the neighborhood corresponding to each feature channel. DCNN applies the diffusion mechanism in the message passing process to extract multi-scale substructure features. To unleash the expressive power of GNN in the graph representation learning, GAT automatically models the connections between nodes by performing a masked self-attention on each node with the shared attentional mechanism to quantify the weights of nodes in the neighborhood aggregation process. GIN proposes an isomorphism GNN framework that injectively propagates information from neighborhoods and achieves maximum discriminative power. Both DCNN and GIN apply sum pooling to generate graph representation, and \cite{xu2018powerful} has shown that the sum pooling takes advantage of the mean/max pooling in distinguishing sets of features. On the other hand, DGCNN and SAGPool develop informative pooling modules to effectively extract the structural information for the classification task. DGCNN sorts concatenated node representations from different GC layers according to nodes' structural role to generate a consistent and meaningful order to sequentially read nodes, while SAGPool hierarchically pools nodes based on node features as well as the graph topology.

\section{Classification Layer Description}
\subsection{Classification}

In the main paper, we have shown that the graph representation after WL-SortPool is grid-structured and is invariant to the order of nodes in input graphs. In SPGNN, we use a standard CNN 
to sequentially read pooled node representations from all layers to learn the embedded subgraph patterns. Let $L$ be the number of GC layers,  the pooled graph representation of WL-SortPool should be $Z_{G} \in \mathbb{R}^{k \times \sum^{L}_{l=1}(d_{l}+1)}$. In analogy to DGCNN, the procedure of generating output to the cross entropy loss is as follows:
\begin{itemize}
  \item We first flatten $Z_{G}$ as a tensor of size $(1, k \times \sum^{L}_{l=1}d_{l})$, and feed this reshaped tensor into a 1-D convolution layer with filter and a step size of $\sum^{L}_{l=1}d_{l}$.  
  \item After that, several 1-D MaxPooling layers and 1-D convolution layers are used to hierarchically extract useful information, and finally, an MLP layer is added to generate the output for the loss construction.
\end{itemize}